\documentclass{article}

\usepackage[final]{neurips_2024}

\usepackage[utf8]{inputenc} 
\usepackage[T1]{fontenc}    
\usepackage{hyperref}       
\usepackage{url}            
\usepackage{booktabs}       
\usepackage{amsfonts}       
\usepackage{nicefrac}       
\usepackage{microtype}      
\usepackage{xcolor}         
\usepackage{bm}
\usepackage{tikz}
\usetikzlibrary{arrows.meta}
\usepackage{subcaption}
\usepackage{calc}
\usepackage{amsmath}
\usepackage{mathtools}

\definecolor{matplotred}{RGB}{201,51,53}
\definecolor{matplotblue}{RGB}{70,124,167}
\definecolor{matplotgreen}{RGB}{88,161,87}
\definecolor{matplotpurple}{RGB}{144,90,152}
\definecolor{matplotorange}{RGB}{223,127,32}

\newcommand{\minisectionNoDot}[1]{\vspace{2mm}\noindent{\textbf{#1}}}
\DeclarePairedDelimiterX{\divbrace}[2]{(}{)}{#1\;\delimsize\|\;#2}
\newcommand{\KLdiv}{D_{\textrm{KL}}\divbrace}

\title{Why Are Parsing Actions\\ for Understanding Message Hierarchies Not Random?}

\author{%
  Daichi~Kato\thanks{This work was conducted while the first author was affiliated with the University of Tokyo.} \\
  Preferred Networks, Inc.\\
  Chiyoda-ku, Tokyo, Japan \\
  \texttt{daichi5967@is.s.u-tokyo.ac.jp} \\
  \AND
  Ryo Ueda \\
  The University of Tokyo \\
  Bunkyo-ku, Tokyo, Japan \\
  \texttt{ryoryoueda@is.s.u-tokyo.ac.jp} \\
  \And
  Yusuke Miyao \\
  The University of Tokyo \\
  Bunkyo-ku, Tokyo, Japan \\
  \texttt{yusuke@is.s.u-tokyo.ac.jp} \\
}

\begin{document}

\maketitle

\begin{abstract}
If humans understood language by randomly selecting parsing actions, it might have been necessary to construct a robust symbolic system capable of being interpreted under any hierarchical structure. However, human parsing strategies do not seem to follow such a random pattern. Why is that the case?
In fact, a previous study on emergent communication using models with hierarchical biases have reported that agents adopting random parsing strategies---ones that deviate significantly from human language comprehension---can achieve high communication accuracy.
In this study, we investigate this issue by making two simple and natural modifications to the experimental setup:
(I) we use more complex inputs that have hierarchical structures, such that random parsing makes semantic interpretation more difficult, and
(II) we incorporate a surprisal-related term, which is known to influence the order of words and characters in natural language, into the objective function.
With these changes, we evaluate whether agents employing random parsing strategies still maintain high communication accuracy.
\end{abstract}

\section{Introduction}

\textbf{Emergent Communication (EC)} \cite{DBLP:journals/corr/Lazaridou_emecom_survey,boldt2024a,Peters2025} is a subfield of computational linguistics that explores the origins of human language through a constructive approach.
One of its main objectives is to analyze and compare human language with the symbol sequences that emerge during learning when neural network-based agents interact while solving cooperative tasks. These emergent symbol systems are regarded as a kind of language---referred to as \textbf{emergent language}.

To induce language emergence among agents, \textbf{Lewis signaling game} \cite{lewis2008convention} is often employed in EC studies \cite{DBLP:conf/nips/ChaabouniKDB19_anti_efficient_encoding,DBLP:conf/acl/Chaabouni_Compo_and_Generalization,DBLP:conf/nips/RitaTMGPDS22,DBLP:conf/iclr/UedaIM23}.\footnote{A variant called \textbf{referential game} or \textbf{discrimination game} is also a popular choice in this field \cite{DBLP:conf/nips/HavrylovT17,DBLP:conf/iclr/LazaridouHTC18_seqRNN_ex1}.}
In the signaling game, there are two types of agents: a \textbf{sender} and a \textbf{receiver}.
The environment defines a \textbf{meaning space} $\mathcal{X}$ and a \textbf{message space} $\mathcal{M}$.
A sample $x \in \mathcal{X}$ is randomly drawn and given only to the sender, who generates a message $m$ based on $x$.
This message $m$ is then passed to the receiver, who uses it to predict the original input $x$ and produces an estimate $\hat{x}$. 

In this formulation, the sender is modeled as $S_\phi (M \mid X)$ using parameters $\phi$, and the receiver is modeled as $R_\theta (X \mid M)$ with parameters $\theta$.
A reward is given to both agents if the predicted $\hat{x}$ matches the original $x$.
Repeating this interaction and training via reinforcement learning allows the agents to develop an emergent language that conveys meaning effectively.

Most EC studies use sequential models such as LSTMs \cite{DBLP:journals/neco/HochreiterS97_LSTM}, GRUs \cite{DBLP:conf/emnlp/ChoMGBBSB14_GRU}, and Transformers \cite{DBLP:conf/nips/VaswaniSPUJGKP17_transformer}.
While these models have achieved considerable practical success in NLP and other domains, they exhibit extremely low data efficiency compared to human language learning \cite{warstadt2023findings_babylm}, raising questions about their suitability as cognitive models of human language acquisition.

In contrast, research outside the EC domain has long assumed that human language involves hierarchical structures such as syntax trees \cite{Chomsky+1957}, and the validity of modeling human language processing as hierarchical continues to be explored \cite{nelson2017neurophysiological}.

Against this backdrop, Kato et al. \cite{kato2024mypaper} investigated how explicitly incorporating a hierarchical bias into the message-processing component of the receiver (i.e., the message encoder) affects EC performance.
They evaluated communication success on unseen data, referred to as \textbf{Communication Accuracy (ComAcc)}, and discussed conditions under which hierarchical models outperform others.
In their experiments, they introduced a baseline called \textbf{random-branching}, in which parsing actions to interpret the hierarchical structure of messages are predicted at random---deviating from human-like strategies---yet surprisingly achieved high ComAcc.

This result may seem counterintuitive at first glance.
However, if the sole objective is to improve communication accuracy, a symbolic system that does not depend on specific hierarchical interpretations could be considered more robust and even desirable.
Still, this does not reflect how actual human parsing strategies work.

Kato et al. suggested that the high performance of random-branching may stem from the simplicity of the meaning space used in their experiments, which allowed successful communication even when message structures were parsed randomly.
They also pointed out that the lack of modeling for surprisal \cite{Levy08}, which is a measure of how unexpected a linguistic unit is given a context and is believed to influence the order of words and characters in human language, might be another contributing factor.

Building on these insights, the present study investigates whether introducing two simple and natural modifications alters the behavior of the random-branching strategy, using a model called \textbf{Stack LSTM} \cite{DBLP:conf/nips/GrefenstetteHSB15, DBLP:journals/corr/abs-1906-01594_Merrill_2018}.
Specifically, we test whether random-branching's ComAcc declines or, even if it does not, whether it leads to results that are implausible as a model of human language understanding\footnote{Due to theoretical requirements in Experiment II \cite{DBLP:conf/iclr/UedaT24_signaling_game_as_vae}, we cannot use the same probabilistic models as previous work, and instead adopt Stack LSTM.
Preliminary experiments (see Appendix \ref{subsec:appendix-reproduce-experiment}) confirmed that Stack LSTM yields results comparable to prior models.
While Experiment I could also be implemented with RL-SPINN, we use Stack LSTM throughout for consistency and simplicity.}.

\minisectionNoDot{Experiment I: Using a More Hierarchically Structured Meaning Space}

Instead of the attribute-value format used by Kato et al. \cite{kato2024mypaper}, we conduct experiments using Dyck-$k$, a context-free language with richer hierarchical structure, as the meaning space.

\minisectionNoDot{Experiment II: Incorporating Surprisal Effects into the Framework}

We conduct experiments using a theoretical formulation that reinterprets the signaling game as a (beta-)VAE \cite{DBLP:journals/corr/KingmaW13, DBLP:conf/iclr/HigginsMPBGBML17}, which naturally incorporates surprisal effects \cite{DBLP:conf/iclr/UedaT24_signaling_game_as_vae}.

In Experiment I, we observe that using a more complex meaning space than attribute-value structures tends to reduce the ComAcc of random-branching.
This suggests that complex meaning spaces may be essential for inducing human-like parsing strategies.
In Experiment II, although modeling surprisal did not alter the trend in ComAcc comparisons, we found that random-branching incurs high cognitive load when interpreting unseen meanings---something typically avoided in human language processing.
This finding supports the view that random-branching is unsuitable as a model of human language understanding.

\section{Background}

\subsection{Stack LSTM}

Stack LSTM is a model that extends a conventional LSTM with a memory structure called a Neural Stack \cite{DBLP:conf/nips/GrefenstetteHSB15}.  
Traditional stacks perform discrete state updates through two operations: \texttt{pop} and \texttt{push}.
Due to the discrete nature of these operations, standard backpropagation cannot be applied, making it generally difficult to incorporate stacks into neural network architectures.

Grefenstette et al. \cite{DBLP:conf/nips/GrefenstetteHSB15} addressed this issue by treating the \texttt{pop} and \texttt{push} operations as continuous values and introducing intermediate stack states, thereby making the entire stack computation differentiable.  
At a given time step $t$, the Neural Stack maintains a vector sequence $\bm{V}_t := (\bm{V}_t [1], \cdots, \bm{V}_t [t])$ and a scalar sequence $\bm{s}_t := (\bm{s}_t [1], \cdots, \bm{s}_t [t])$.  
Here, $\bm{V}_t[i]$ denotes the $i$-th element in the stack (counting from the bottom), and $\bm{s}_t[i]$ represents the “strength” or the degree to which $\bm{V}_t[i]$ remains in the stack.

At each time step $t$, the Neural Stack receives as input the previous stack state $\bm{V}_{t-1}, \bm{s}_{t-1}$, a vector $\bm{v}_t$ to be pushed onto the stack, and scalar values $u_t$, $d_t$, and $r_t$ indicating the intensities of the \texttt{pop}, \texttt{push}, and \texttt{read} operations, respectively.  
Within a single time step, the Neural Stack performs the \texttt{pop}, \texttt{push}, and \texttt{read} operations in that order.  
The \texttt{pop} operation decreases the strength of the topmost element by $u_t$, recursively removing strength from lower elements if the top one is insufficient.  
The \texttt{push} operation adds $\bm{v}_t$ to the top of the stack with strength $d_t$.  
The \texttt{read} operation, similar to \texttt{pop}, traverses the stack from the top until accumulating strength $r_t$, computing a weighted sum of elements, which becomes the read vector $\bm{r}_t$ at time $t$.

Furthermore, Grefenstette et al. \cite{DBLP:conf/nips/GrefenstetteHSB15} proposed the \textbf{Stack RNN} model, which combines the Neural Stack with a controller---a recurrent neural network that outputs inputs to the Neural Stack.  
In particular, when the controller is an LSTM, the resulting architecture is referred to as a \textbf{Stack LSTM}.
In our experiments, we employ a variant of Stack LSTM proposed by Merrill et al. \cite{DBLP:journals/corr/abs-1906-01594_Merrill_2018}, which allows the \texttt{pop}, \texttt{push}, and \texttt{read} values to exceed 1.

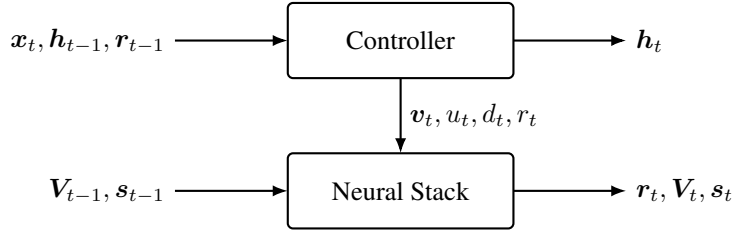
\begin{figure}[t]
\centering
\begin{tikzpicture}[scale=1.0]
    \draw[thick, rounded corners=2pt](-1.5,2)--(-1.5,1)--(1.5,1)--(1.5,2)-- cycle;
    \node at (0, 1.5) {Controller};
    \draw[-{Latex[length=2mm]}, thick] (-3, 1.5) node[anchor=east]{$\bm{x}_t, \bm{h}_{t-1}, \bm{r}_{t-1}$}--(-1.5, 1.5);
    \draw[-{Latex[length=2mm]}, thick] (1.5, 1.5)--(3, 1.5) node[anchor=west]{$\bm{h}_t$};

    \draw[-{Latex[length=2mm]}, thick] (0, 1)--(0, 0);
    \node[anchor=west] at (0, 0.5) {$\bm{v}_t, u_t, d_t, r_t$};

    \draw[thick, rounded corners=2pt](-1.5,-1)--(-1.5,0)--(1.5,0)--(1.5,-1)-- cycle;
    \node at (0, -0.5) {Neural Stack};
    \draw[-{Latex[length=2mm]}, thick] (-3, -0.5) node[anchor=east]{$\bm{V}_{t-1}, \bm{s}_{t-1}$}--(-1.5, -0.5);
    \draw[-{Latex[length=2mm]}, thick] (1.5, -0.5)--(3, -0.5) node[anchor=west]{$\bm{r}_t, \bm{V}_t, \bm{s}_t$};
\end{tikzpicture}
\caption{Overview of the Stack RNN.}\label{fig:stack-rnn}
\end{figure}

\subsection{A Framework for Reinterpreting Signaling Games as VAEs}\label{subsec:シグナリングゲームをVAEとして再解釈する枠組み}

Ueda et al. \cite{DBLP:conf/iclr/UedaT24_signaling_game_as_vae} observed that the roles of the sender and receiver in signaling games closely correspond to those of the encoder and decoder in VAEs.  
They proposed redefining the receiver as the joint distribution over meanings and messages:
\begin{equation*}
    R_\theta^{\mathrm{joint}} (X, M) := P_\theta^{\mathrm{prior}} (M) R_\theta (X \mid M)
\end{equation*}
where $P_\theta^{\mathrm{prior}} (M)$ is the prior distribution over messages.  

Inspired by the objective function of $\beta$-VAE \cite{DBLP:conf/iclr/HigginsMPBGBML17}, they redefined the signaling game's objective $\mathcal{J}_{\mathrm{ec-vae}}$ as follows\footnote{Ueda et al. \cite{DBLP:conf/iclr/UedaT24_signaling_game_as_vae} applied an annealing schedule to gradually increase the weight $\beta$ toward 1, using REWO \cite{DBLP:conf/nips/KlushynCKCS19}. Our experiments adopt the same setting.}:
\begin{equation*}
        \mathcal{J}_{\mathrm{ec-vae}} = \mathbb{E}_{x \sim P_{\mathrm{inp}}(X)} \left[\vphantom{P_\theta^{\mathrm{prior}}}\right.
            \mathbb{E}_{m \sim S_\phi (M \mid x)} \left[\log R_\theta (x \mid m)\right]
            - \beta \KLdiv{S_\phi (M \mid x)}{P_\theta^{\mathrm{prior}} (M)}
        \left.\vphantom{P_\theta^{\mathrm{prior}}}\right]
\end{equation*}

They showed that by reformulating the objective, a natural trade-off emerges between reconstructing the meaning from the message and minimizing surprisal, both of which are implicitly captured by the objective function.

\section{Experimental Method}

In this section, we provide a concise overview of our experimental methodology.  
For full details, please refer to Appendix~\ref{subsec:appendix-hyperparameter}.

\subsection{Experiment I}

\paragraph{Experimental Setup:}
Our setup largely follows the prior work by Kato et al.~\cite{kato2024mypaper}, with two primary modifications:  
(1) we replace the message encoder, which previously used RL-SPINN, with a Stack LSTM~\cite{DBLP:journals/corr/abs-1906-01594_Merrill_2018};  
(2) we use Dyck-$k$ as the meaning space.  

Dyck-$k$ is a context-free language consisting of strings with properly nested parentheses of $k$ different types. It is formally defined by the following context-free grammar:
\begin{equation*}
    S \to (_i \; S \; )_i \; S \mid \varepsilon \quad (1 \leq i \leq k)
\end{equation*}

To constrain the size of the meaning space, we limit the sentence length using a maximum length parameter $l_{\mathrm{max}}$.  
The tested configurations for the meaning space are: $(k, l_{\mathrm{max}}) = (1, 18), (4, 8), (9, 6)$.

\paragraph{Baselines:}
In addition to our Stack LSTM-based model, which explicitly incorporates a structural bias toward hierarchy, we implement two baseline models: \textbf{left-branching} and \textbf{random-branching}, following the conventions in prior work~\cite{kato2024mypaper}.  
To ensure fair comparison, the only difference between models lies in how the strengths for \texttt{pop}, \texttt{push}, and \texttt{read} operations are determined.

In our model, the LSTM controller freely determines the strengths within the ranges $u_t \in [0, k_u]$, $d_t \in [0, k_d]$, and $r_t \in [0, k_r]$.  
In contrast, the \textbf{left-branching} baseline fixes $u_t = 1$, $d_t = 1$, and $r_t = 1$ to simulate a completely left-branching structure---analogous to the sequential order in standard models.  
The \textbf{random-branching} baseline samples the strengths independently from a uniform distribution over the same ranges as the controller.  
Note that even for the same input message, the strength values are randomly sampled each time.

\subsection{Experiment II}

\paragraph{Experimental Setup:}
The basic setup follows that of Experiment I.  
As described in Section~\ref{subsec:シグナリングゲームをVAEとして再解釈する枠組み}, the receiver is tasked not only with reconstructing the meaning but also with modeling the prior distribution over messages.

To achieve this, we augment the receiver with an architecture that, at each time step $t$, predicts the conditional prior distribution $P_\theta^{\mathrm{prior}} (M_t \mid M_{1:t-1})$ using the previous read vector $\bm{r}_{t-1}$.  
After reaching the final time step, we compute the overall message probability as:
\[
P_\theta^{\mathrm{prior}} (M) = \prod_{t=1}^{|M|} P_\theta^{\mathrm{prior}} (M_t \mid M_{1:t-1})
\]

Following Kato et al.~\cite{kato2024mypaper}, the meaning space is defined using an attribute-value representation.  
We test two configurations: $(n_{\mathrm{att}}, n_{\mathrm{val}}) = (2, 64)$ and $(4, 8)$,  
where $n_{\mathrm{att}}$ denotes the number of attributes, and $n_{\mathrm{val}}$ is the number of possible values for each attribute.

\section{Experimental Results and Discussion}

All experiments were conducted using 24 different random seeds for each setting. The aggregated results are shown in \autoref{fig:results}.  
For Experiment II, we excluded runs in which the $\beta$ coefficient for the KL divergence did not reach 0.95 or higher by the final iteration.

\newcommand{\widthPerImage}{0.14}
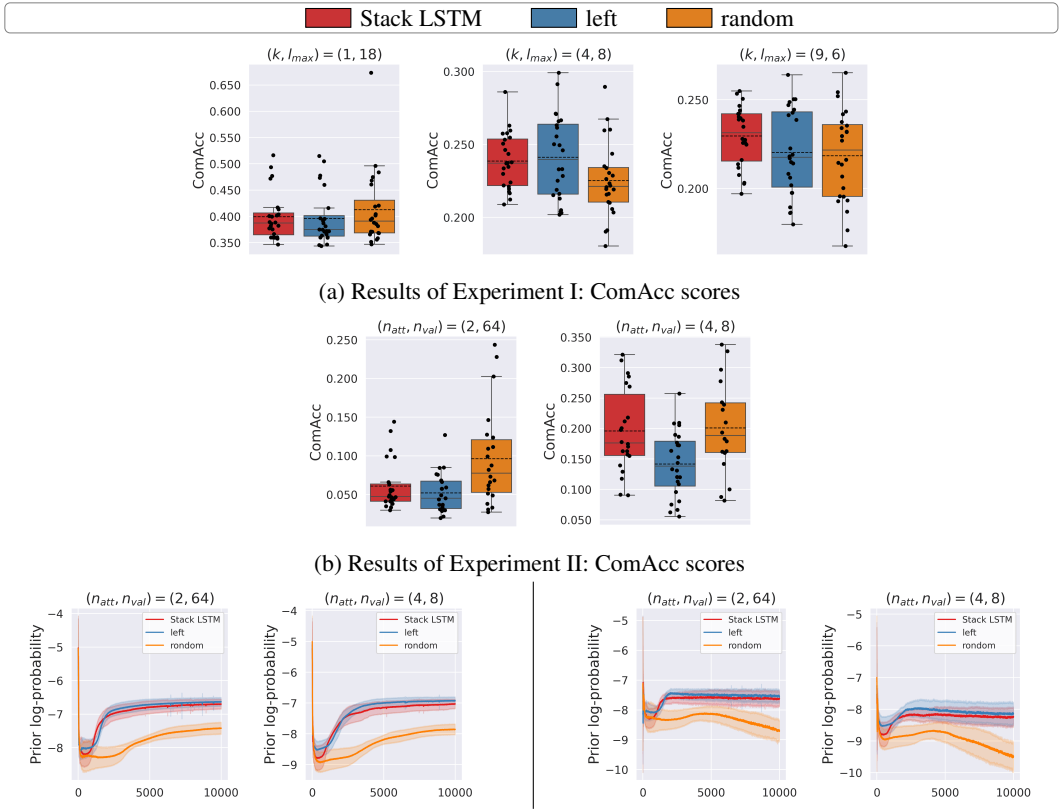
\begin{figure*}[t]
    \centering
    \begin{minipage}{\linewidth}
        \centering
        \begin{tikzpicture}
            \draw [very thin, color=gray, rounded corners=2pt] (-3.0,0)--(11,0)--(11,0.4)--(-3.0,0.4)-- cycle;
            \draw [fill=matplotred] (1, 0.325) rectangle node [anchor=west, xshift=0.3cm] {\small Stack LSTM} (1.6, 0.075);
            \draw [fill=matplotblue] (4, 0.325) rectangle node [anchor=west, xshift=0.3cm] {\small left} (4.6, 0.075);
            \draw [fill=matplotorange] (5.8, 0.325) rectangle node [anchor=west, xshift=0.3cm] {\small random} (6.4, 0.075);
        \end{tikzpicture}
    \end{minipage}
    \begin{minipage}{\linewidth}
        \centering
        \includegraphics[width=\widthPerImage\paperwidth]{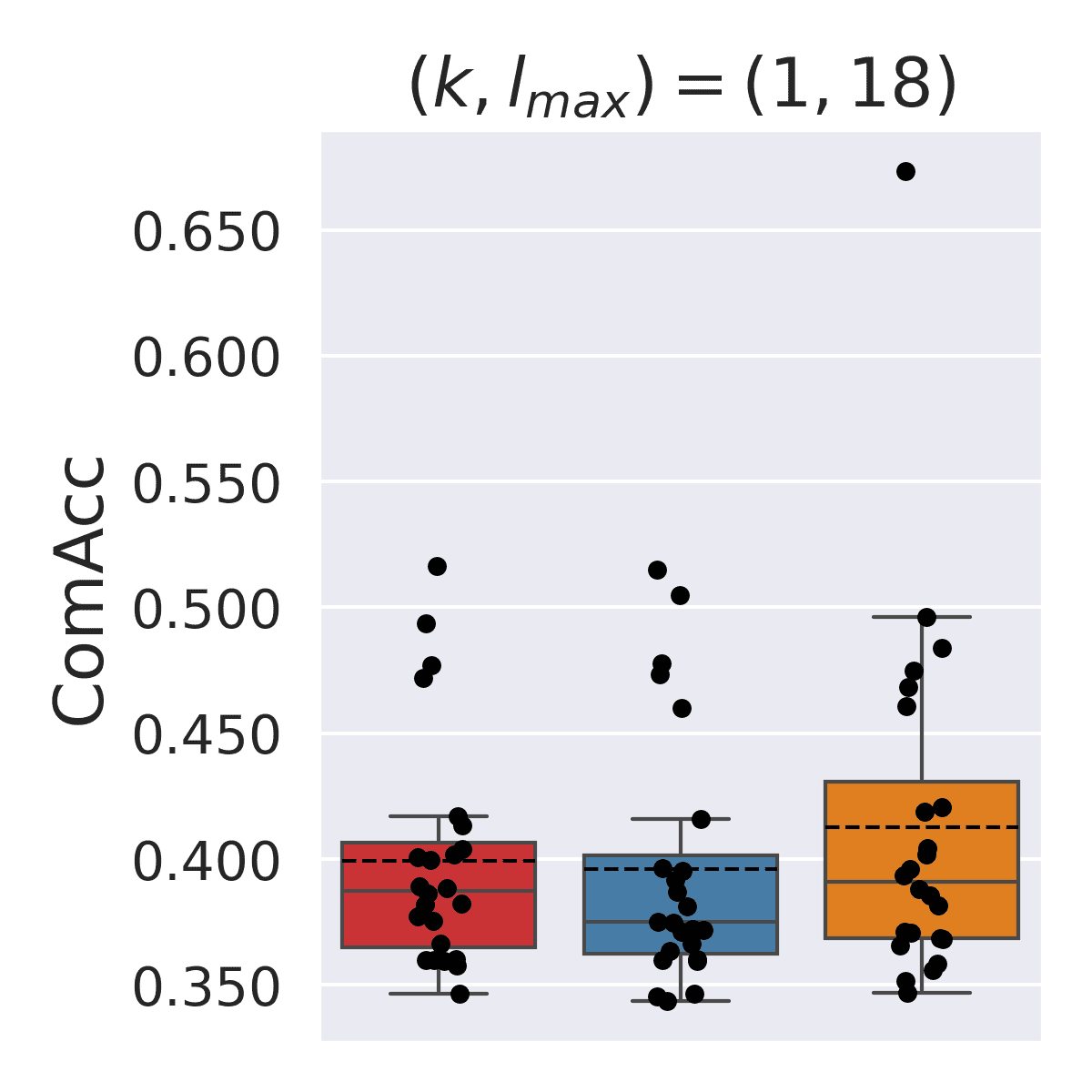}
        \includegraphics[width=\widthPerImage\paperwidth]{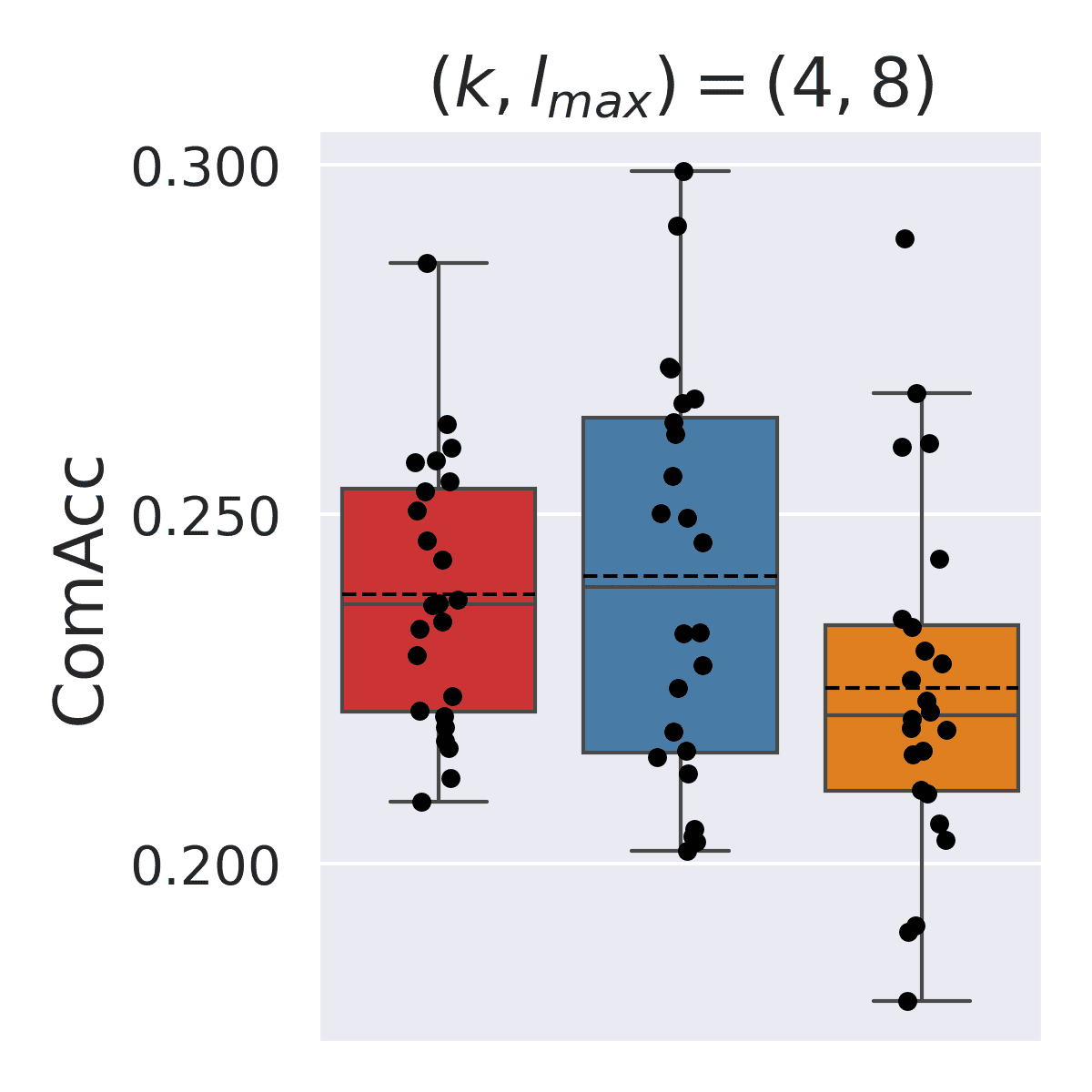}
        \includegraphics[width=\widthPerImage\paperwidth]{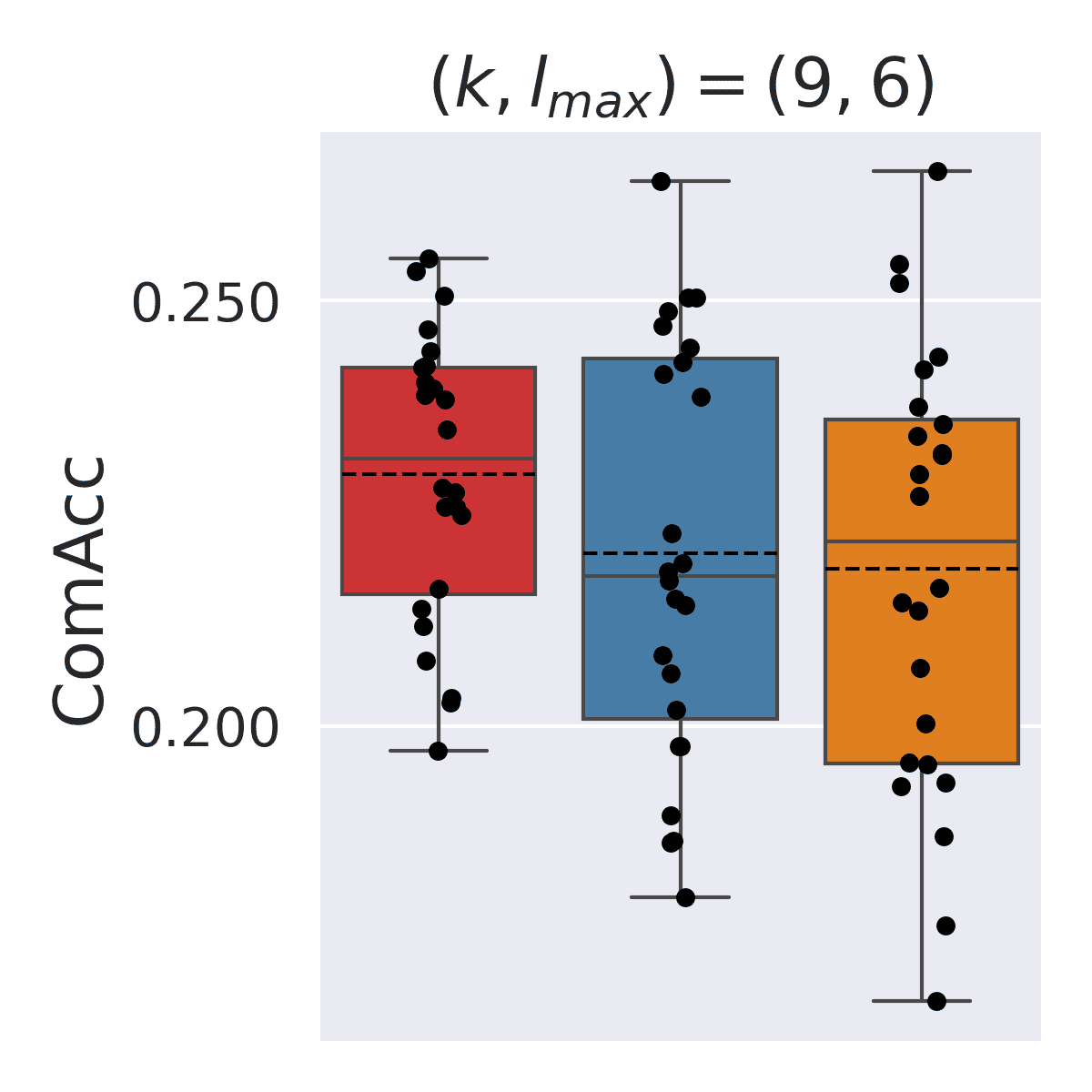}
        \subcaption{Results of Experiment I: ComAcc scores}\label{fig:results-exp1}
    \end{minipage}
    \begin{minipage}{\linewidth}
        \centering
        \includegraphics[width=\widthPerImage\paperwidth]{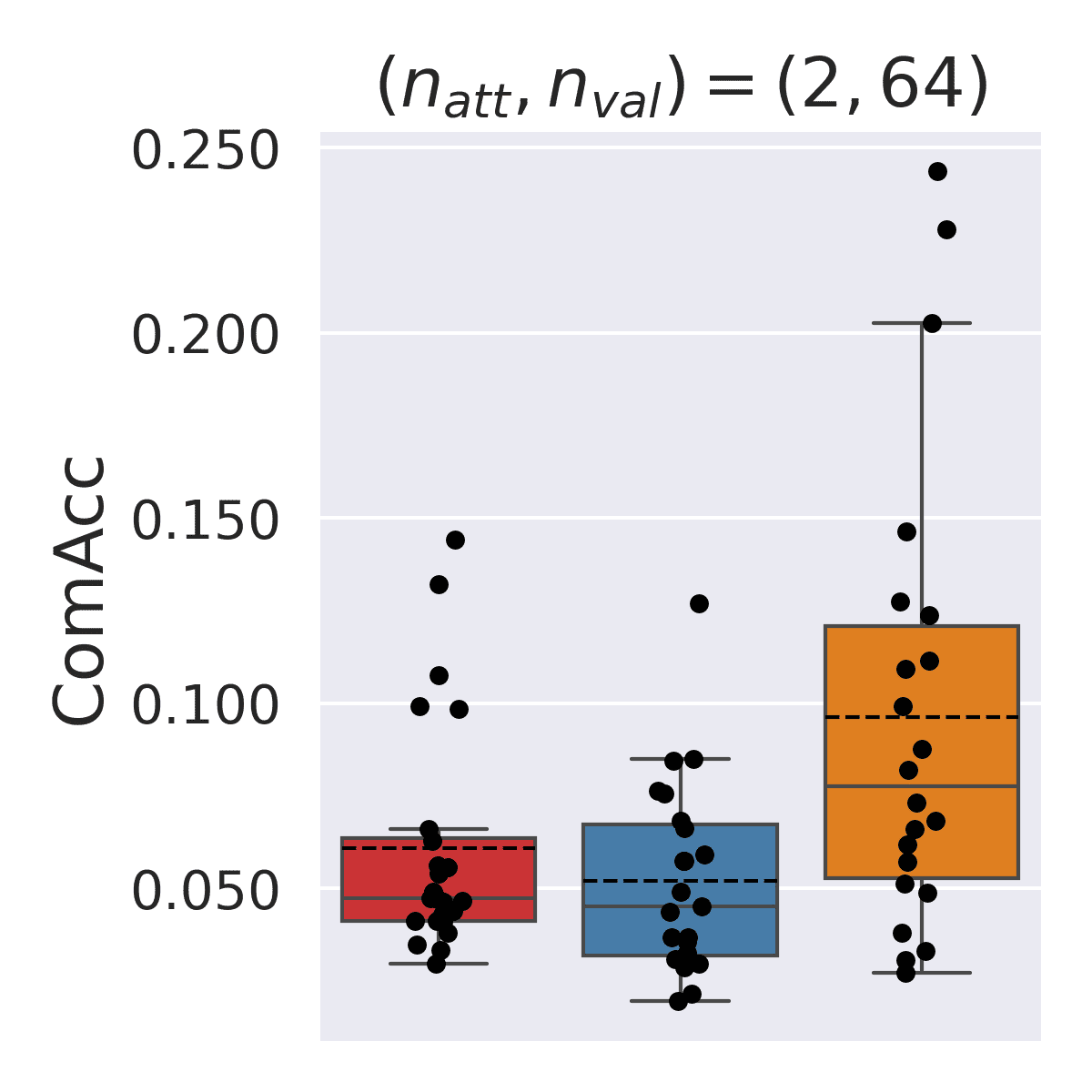}
        \includegraphics[width=\widthPerImage\paperwidth]{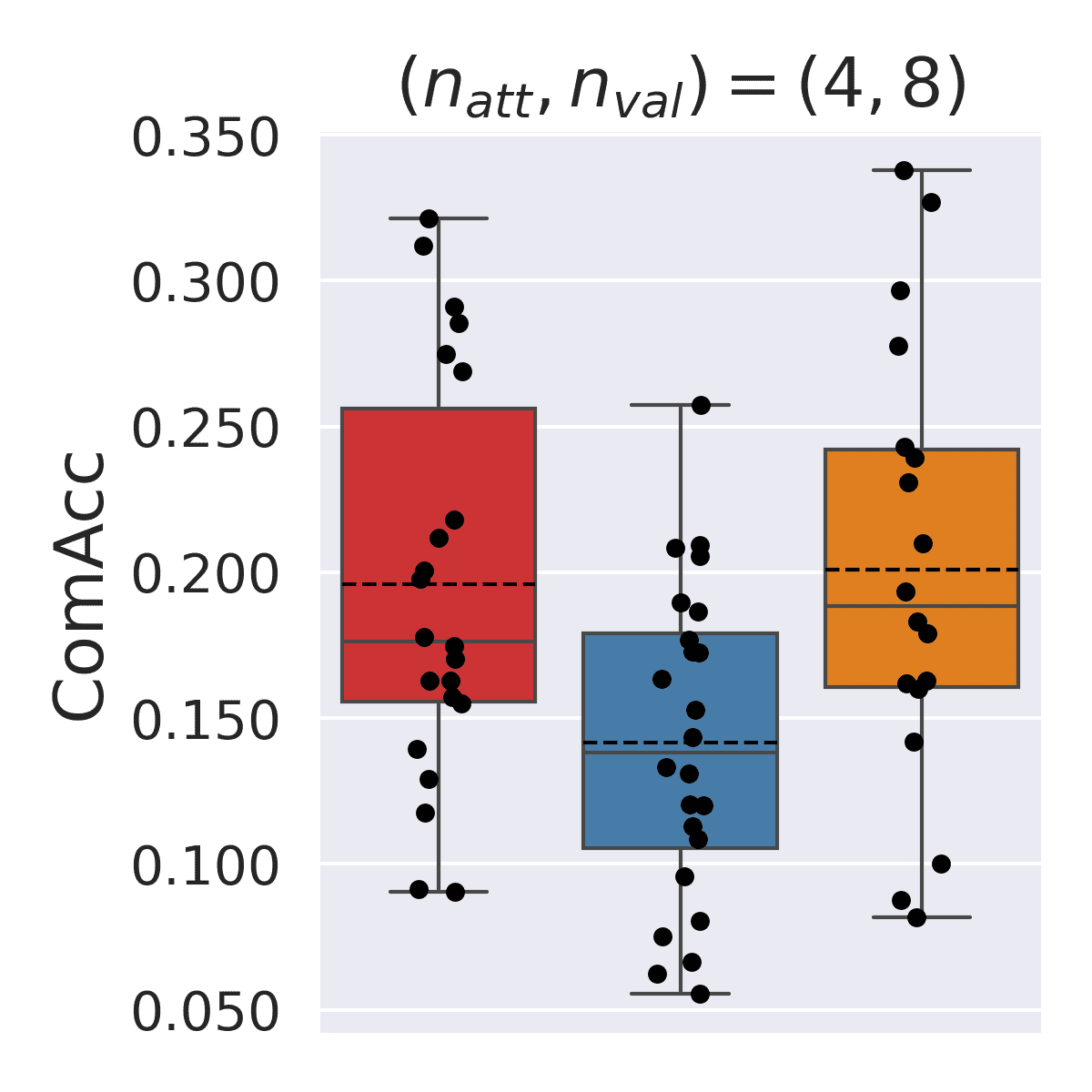}
        \subcaption{Results of Experiment II: ComAcc scores}\label{fig:results-exp2-comacc}
    \end{minipage}
    \begin{minipage}{\linewidth}
        \centering
        \includegraphics[width=\widthPerImage\paperwidth]{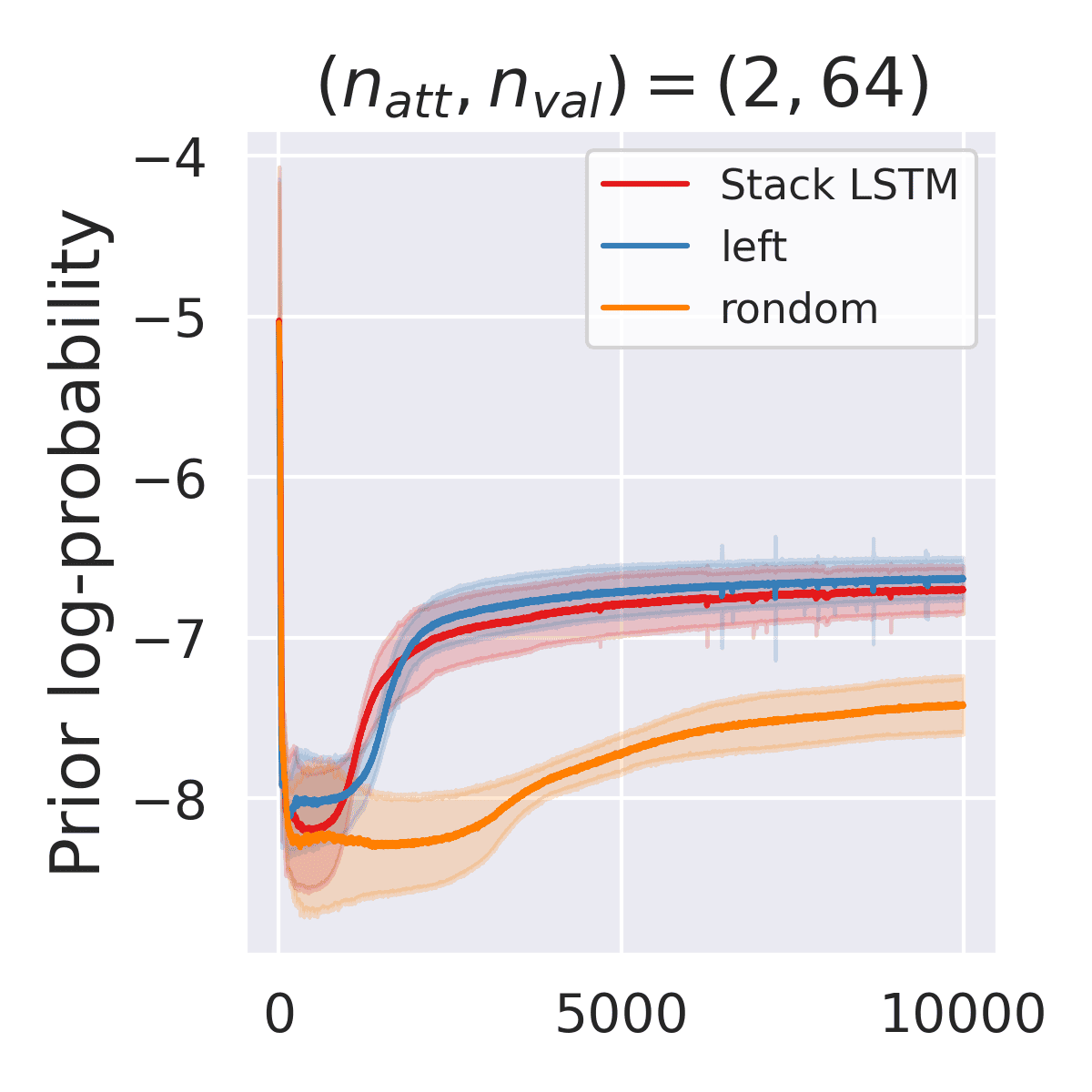}
        \includegraphics[width=\widthPerImage\paperwidth]{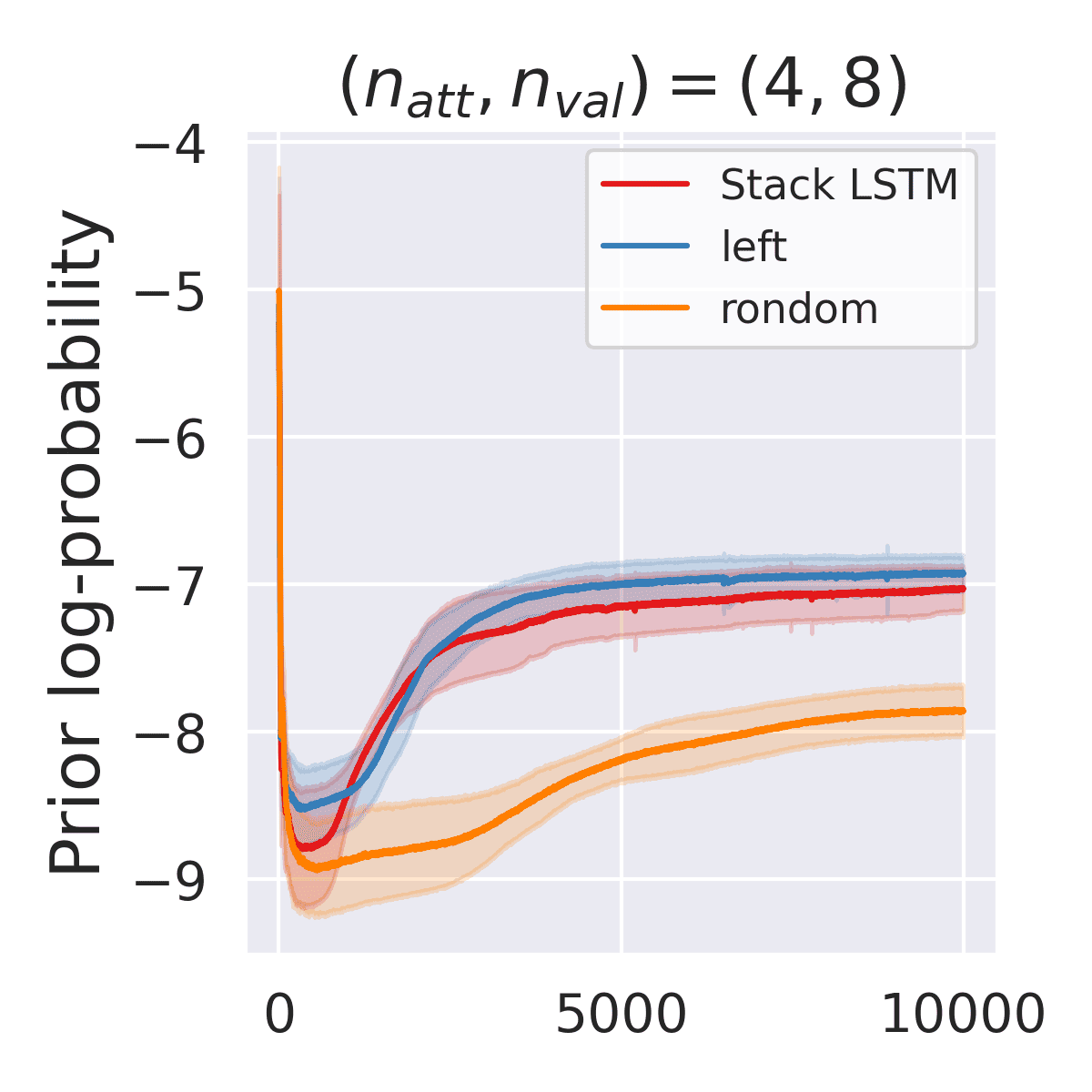}
        \hspace{0.5cm}
        \vline
        \hspace{0.5cm}
        \includegraphics[width=\widthPerImage\paperwidth]{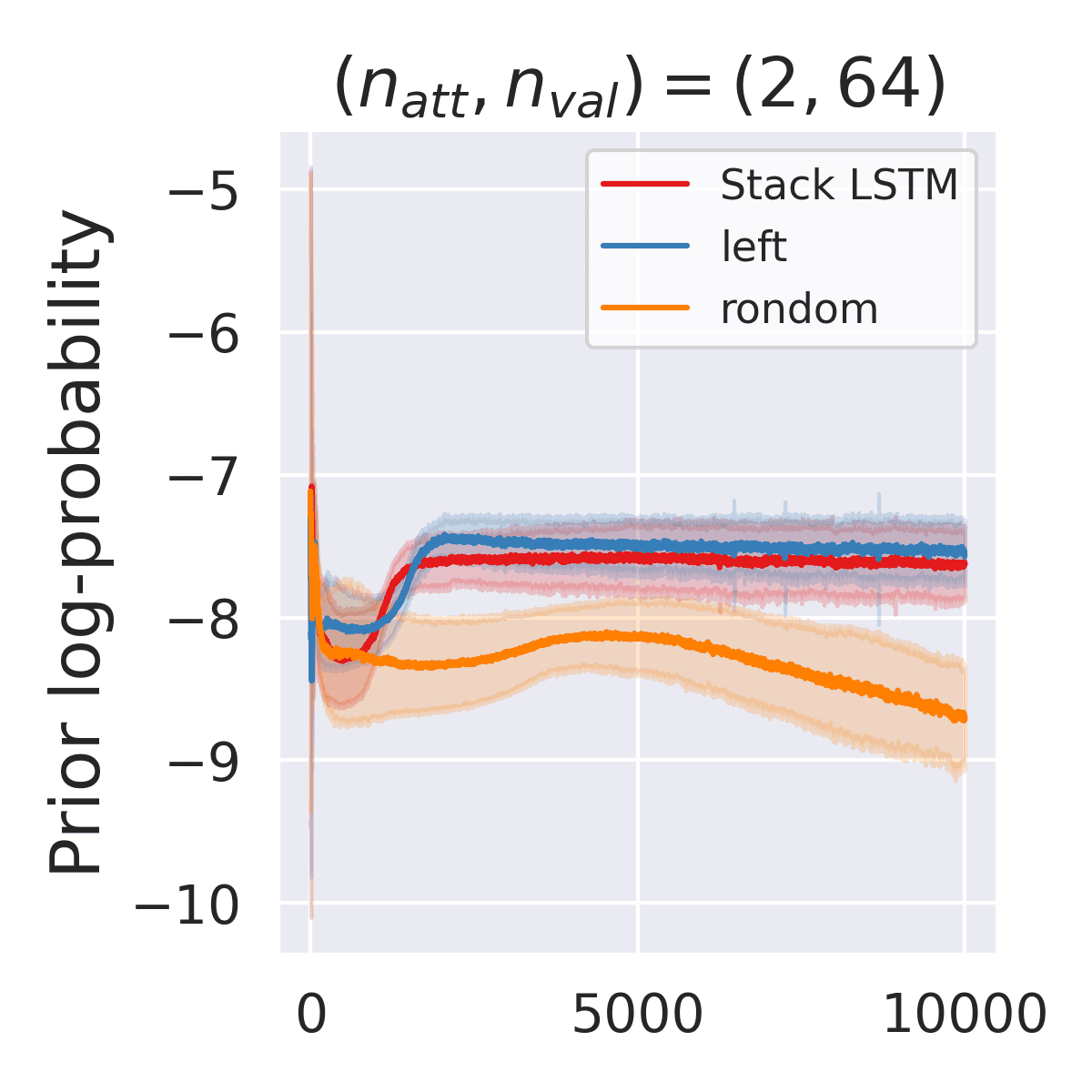}
        \includegraphics[width=\widthPerImage\paperwidth]{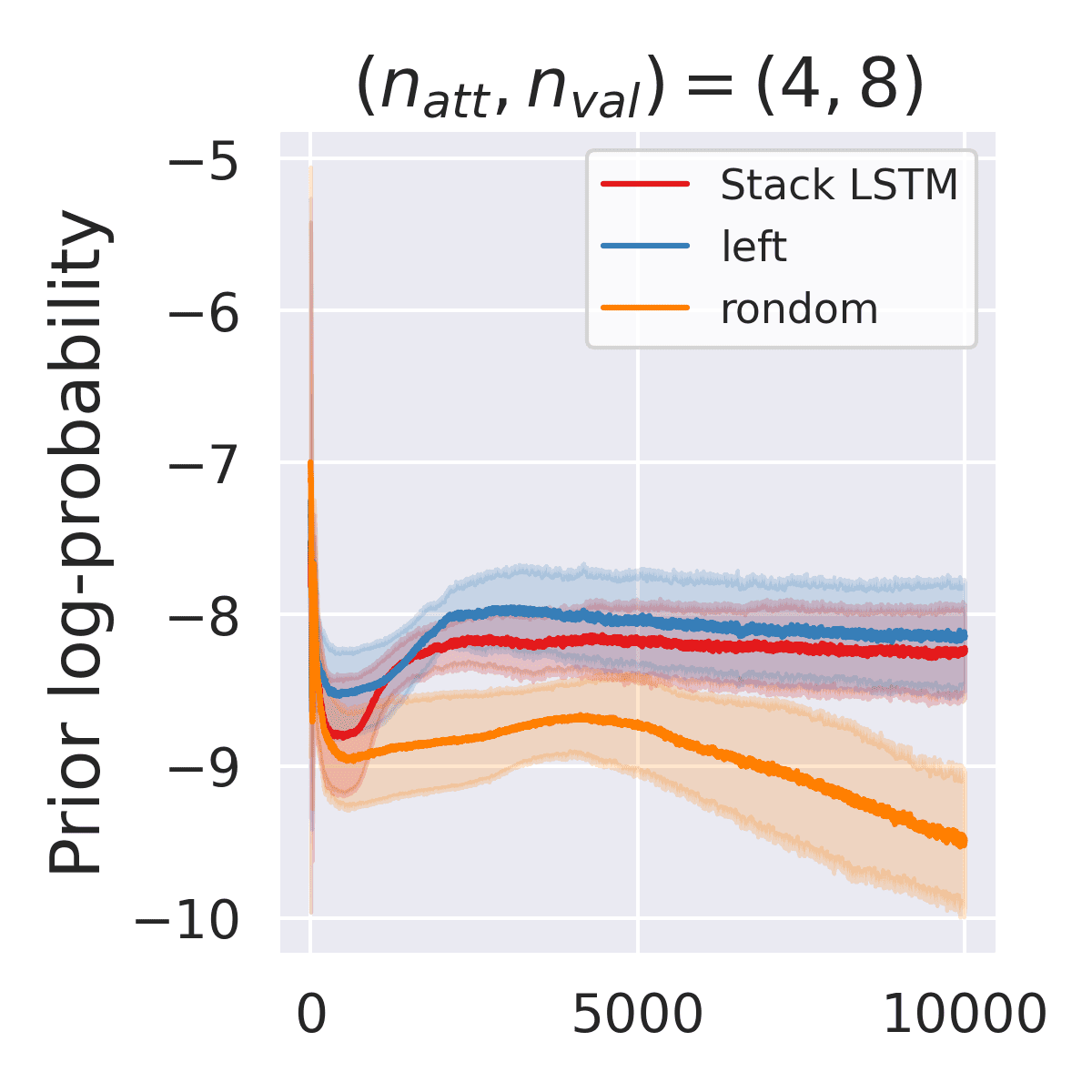}
        \subcaption{Results of Experiment II for $\log P_\theta^{\mathrm{prior}} (M)$. The horizontal axis indicates the number of iterations, and the vertical axis shows the log-probability of the message. The left two plots represent training data results, while the right two show test data results.}\label{fig:results-exp2-log-probs}
    \end{minipage}
    \caption{Experimental results}\label{fig:results}
\end{figure*}

\subsection{Experiment I}

As shown in \autoref{fig:results-exp1}, in settings with larger values of $k$, the ComAcc scores for the random-branching models tend to be lower than those of other models.  
In Dyck-$k$ tasks with $k=1$, it suffices to track only the nesting depth---i.e., the number of open parentheses at each timestep---so the semantic space is essentially non-hierarchical and can be handled reasonably well even by the random-branching models.  
In contrast, when $k > 1$, it becomes necessary to precisely remember not only the nesting depth but also the order in which different types of parentheses were opened.  
This poses a challenge for random-branching models, which have difficulty preserving accurate hierarchical message structures, resulting in reduced ComAcc scores.

These findings suggest that in EC experiments, using a more structurally complex semantic space---beyond attribute-value representations---can be beneficial.

\subsection{Experiment II}

As seen in \autoref{fig:results-exp2-comacc}, introducing a surprisal-related term did not significantly affect the overall trend that the ComAcc of the random-branching condition tends to be higher.

However, as shown in \autoref{fig:results-exp2-log-probs}, in our Stack LSTM-based model as well as the left-branching model, the predictability of messages during training tends to improve along with their predictability on the test set.  
In contrast, the random-branching model exhibits a kind of overfitting: as message predictability improves on the training data, it deteriorates on the test data.

From the perspective of surprisal, when attempting to interpret unseen meanings, receivers using random-branching structures interpret messages with greater ``surprise.''  
Such models, which are constantly subjected to high cognitive costs, cannot be considered human-like.  
This viewpoint offers one possible explanation for the unsuitability of using random-branching structures in EC.

\section{Conclusion}

In this experiment, we empirically tested the interpretation presented by Kato et al. \cite{kato2024mypaper}, which suggested that the ComAcc of agents with randomness in predicting the hierarchical structure of messages tends to be undesirably high.

In Experiment I, we employed a more complex semantic space with hierarchical structure, rather than an attribute-value representation, and observed a tendency for the ComAcc gap between random agents and other models to diminish.  
This finding suggests the potential significance of using hierarchically structured semantic spaces in the context of EC.

In Experiment II, we incorporated a surprisal-based component into the environment and confirmed that a random agent experienced high cognitive load---typically avoided in human communication---when interpreting messages corresponding to unseen meanings.  
This result may serve as empirical support for the claim that random agents are inappropriate as models of human communication behavior.

\begin{ack}
This work was supported by 
JSPS KAKENHI Grant Number JP23KJ0768 
and JST ACT-X Grant Number JPMJAX24C5. 
\end{ack}

\bibliographystyle{unsrt}
\bibliography{myref}


\appendix

\newpage

\section{Experimental Details}\label{subsec:appendix-hyperparameter}

\subsection{Experiment I}\label{subsubsec:appendix-hyperparameter-exp1}

Regarding the message space, the maximum message length is set to 8, and the number of symbols, including EOS, is 4.  
For the architecture parameters, both the sender and receiver use 512-dimensional hidden vectors, and the vectors stored in the Neural Stack are also 512-dimensional. The embedding vectors for the symbols are 32-dimensional.  
The maximum strengths of the \texttt{pop}, \texttt{push}, and \texttt{read} operations, denoted by $k_u$, $k_d$, and $k_r$, respectively, are all set to 2.  

We use the Adam optimizer \cite{DBLP:journals/corr/KingmaB14_Adam} with both the learning rate and the L2 regularization coefficient set to 0.0001.  
The coefficient for the entropy regularization term applied to the sender is set to 0.5.  

The meaning space is split in a 9:1 ratio, with the larger portion used for training and the smaller for testing.  
Each training iteration uses a batch size of 8192, and the model is trained for 15,000 iterations per run.

\subsection{Experiment II}\label{subsubsec:appendix-hyperparameter-exp2}

We use the same settings as in \autoref{subsubsec:appendix-hyperparameter-exp1}, unless otherwise noted.  
The initial value of the coefficient $\beta$ for the KL divergence in the REWO algorithm \cite{DBLP:conf/nips/KlushynCKCS19} is set to 0.001.  
The model is trained for 10,000 iterations per run.

\section{Preliminary Experiment: Reproduction of Prior Work Using Stack LSTM \texorpdfstring{\cite{kato2024mypaper}}{}}\label{subsec:appendix-reproduce-experiment}

\begin{figure}[t]
    \centering
    \begin{minipage}{\linewidth}
        \centering
        \begin{tikzpicture}
            \draw [very thin, color=gray, rounded corners=2pt] (-3.0,0)--(11,0)--(11,0.4)--(-3.0,0.4)-- cycle;
            \draw [fill=matplotred] (1, 0.325) rectangle node [anchor=west, xshift=0.3cm] {\small Stack LSTM} (1.6, 0.075);
            \draw [fill=matplotblue] (4, 0.325) rectangle node [anchor=west, xshift=0.3cm] {\small left} (4.6, 0.075);
            \draw [fill=matplotorange] (5.8, 0.325) rectangle node [anchor=west, xshift=0.3cm] {\small random} (6.4, 0.075);
        \end{tikzpicture}
    \end{minipage}
    \begin{minipage}{\linewidth}
        \centering
        \includegraphics[width=\widthPerImage\paperwidth]{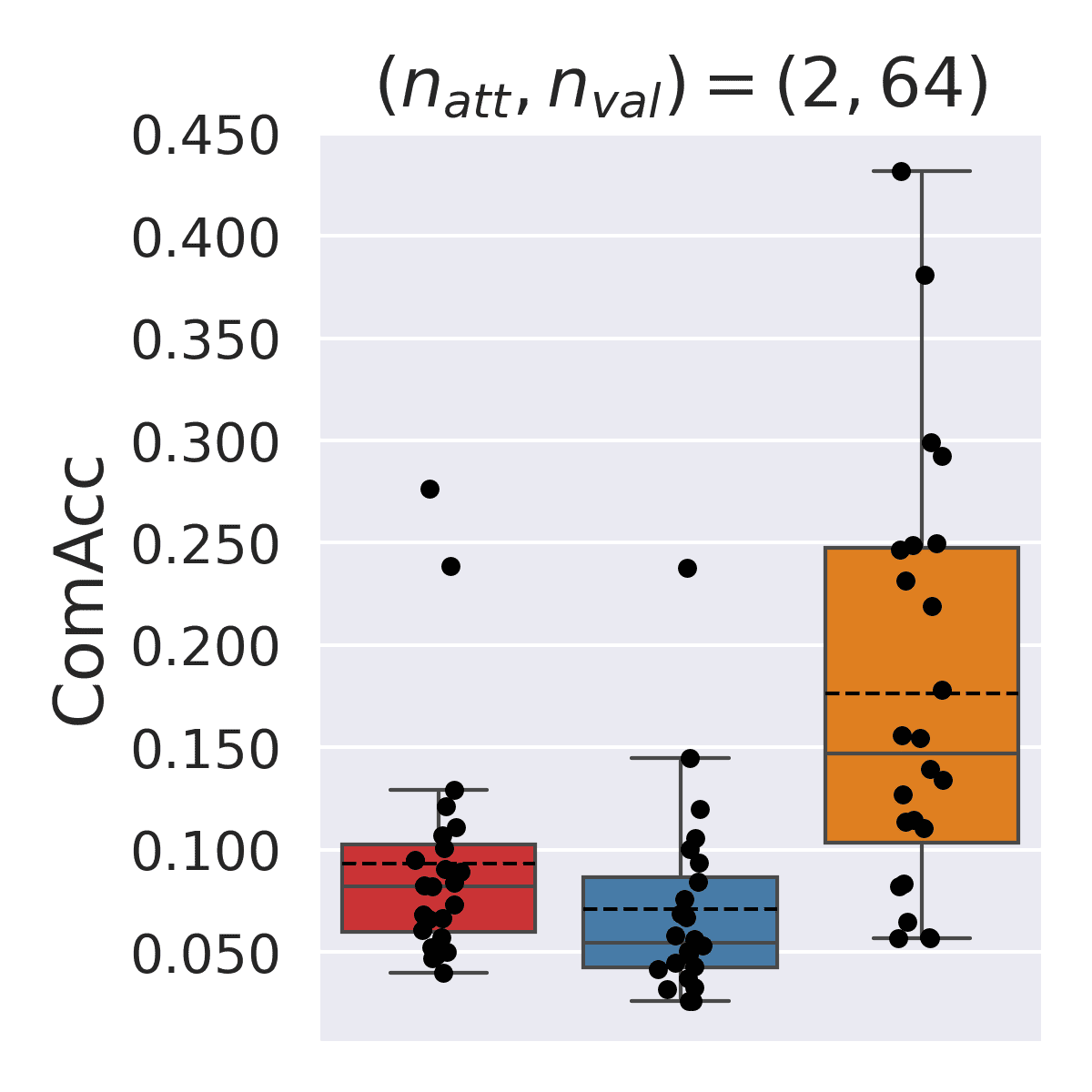}
        \includegraphics[width=\widthPerImage\paperwidth]{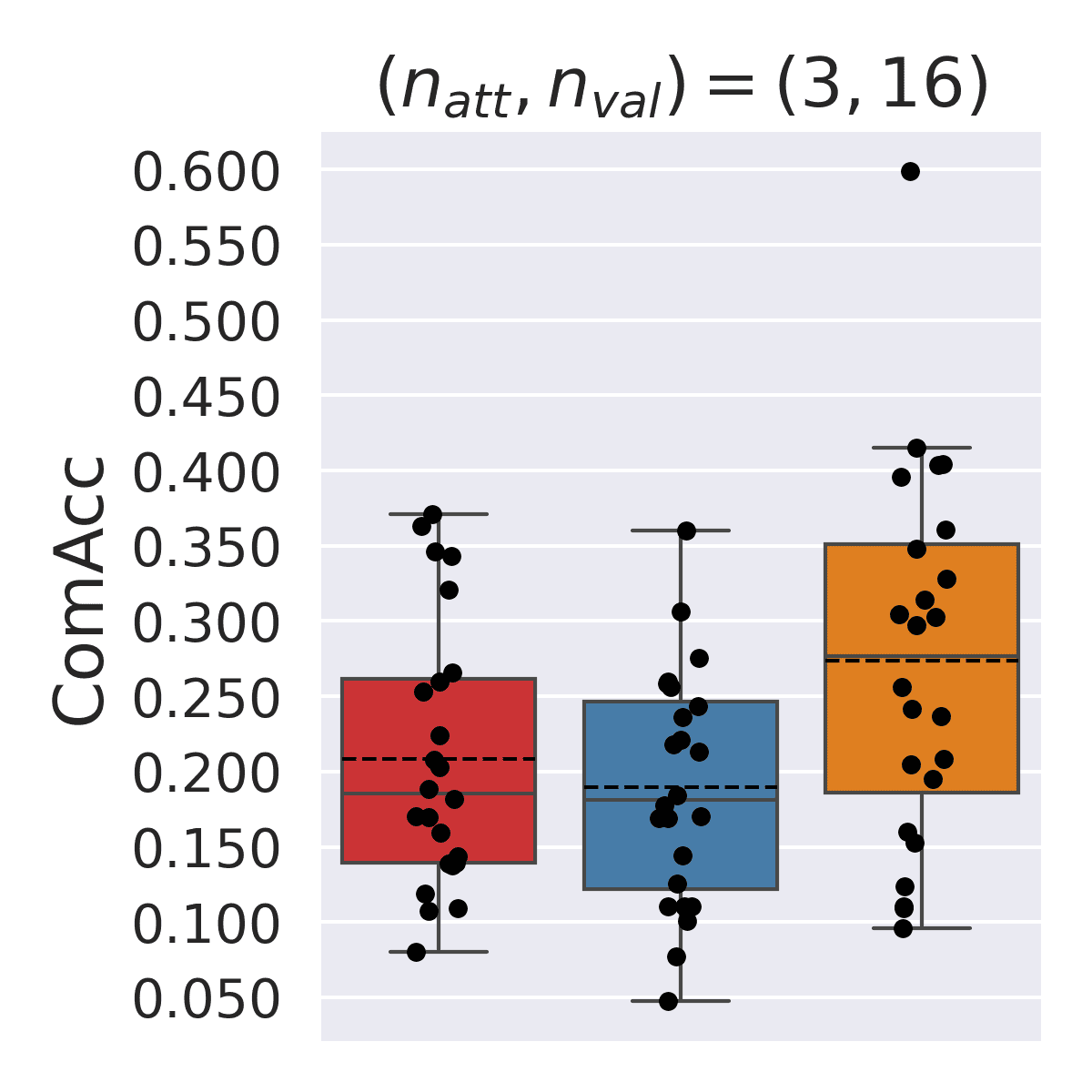}
        \includegraphics[width=\widthPerImage\paperwidth]{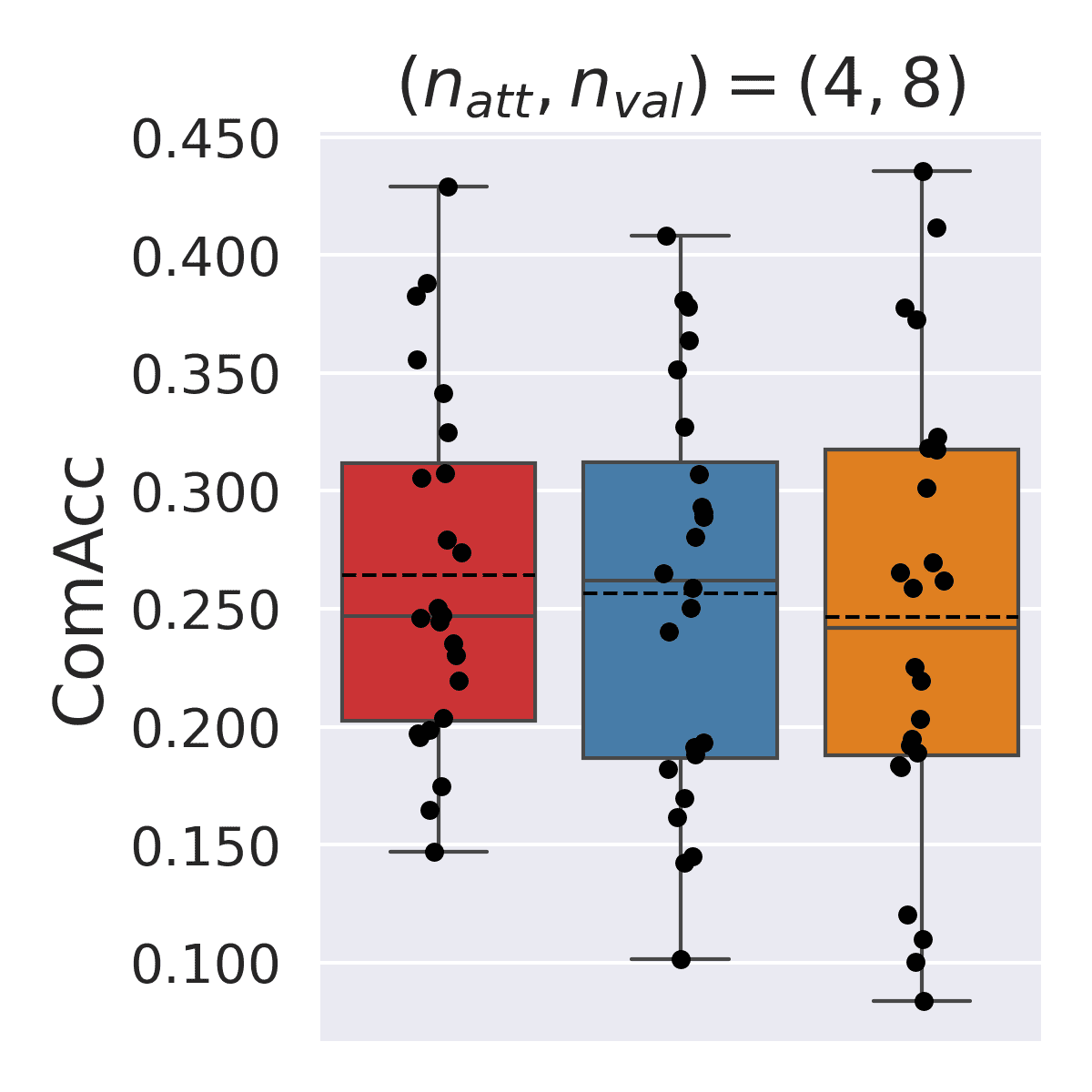}
        \includegraphics[width=\widthPerImage\paperwidth]{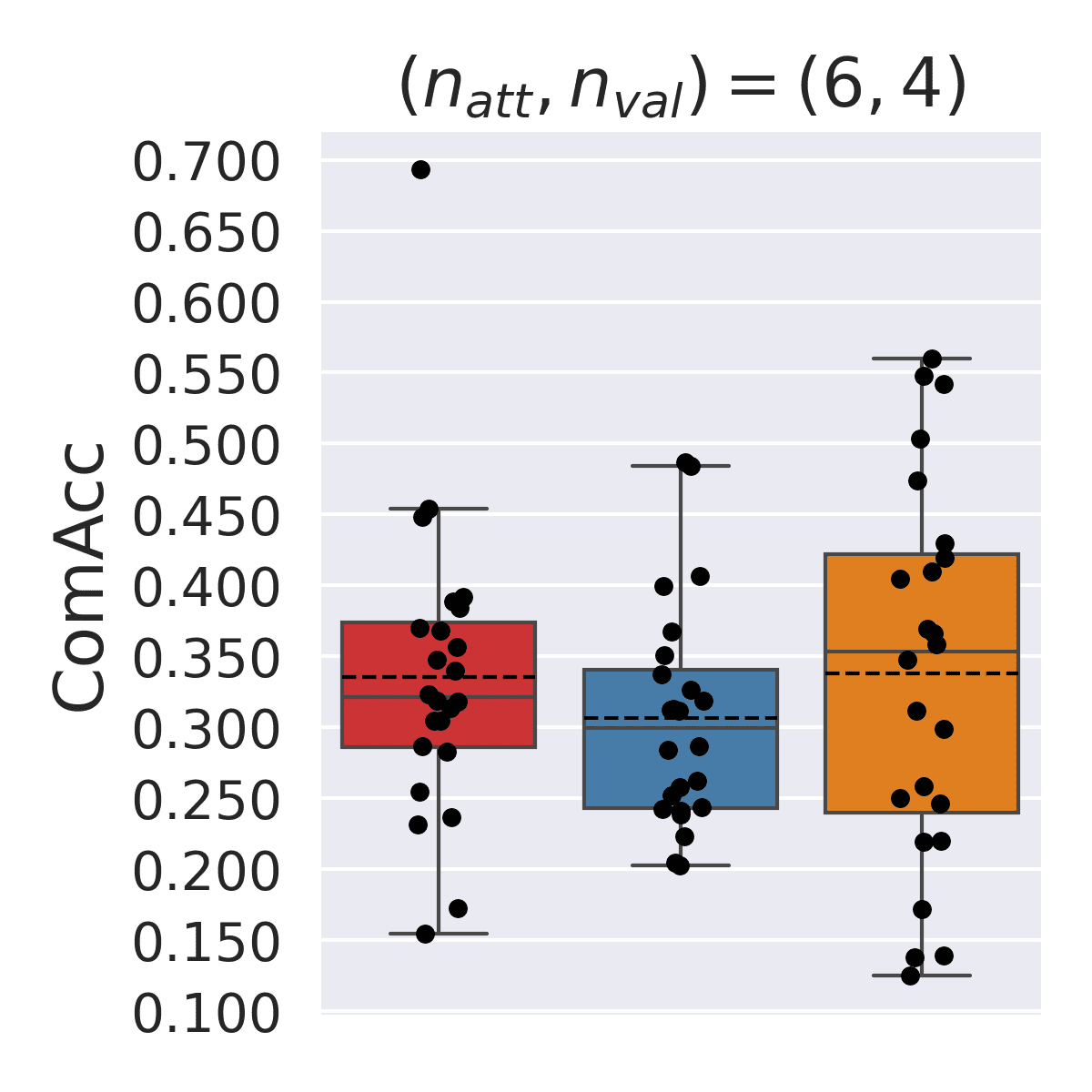}
    \end{minipage}
    \caption{ComAcc results of the preliminary experiment}\label{fig:results-exp-pre}
\end{figure}
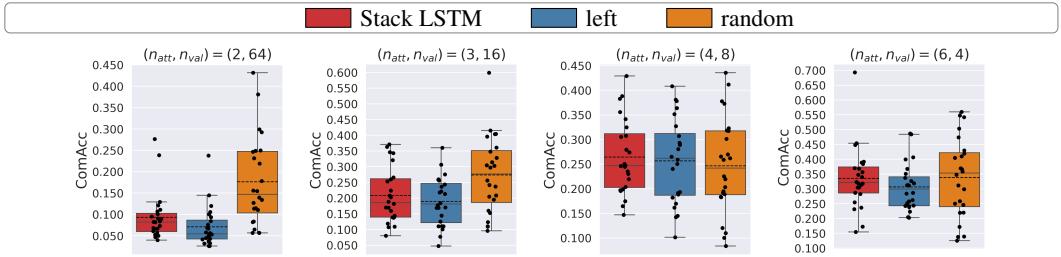

\subsection{Experimental Setup}

We adopt the same settings as in \autoref{subsubsec:appendix-hyperparameter-exp1}.  
The meaning space configurations tested are $(n_{\mathrm{att}}, n_{\mathrm{val}}) = (2, 64), (3, 16), (4, 8), (6, 4)$.  
Each run consists of 5,000 training iterations.

\subsection{Experimental Results}

As shown in \autoref{fig:results-exp-pre}, we observed results similar to those reported in the prior work \cite{kato2024mypaper}, where the random-branching condition yields higher ComAcc scores.  
This suggests that the phenomenon of increased ComAcc under random branching is not specific to a particular model, but rather that the randomness itself contributes positively to ComAcc performance.

 
\end{document}